\documentclass[5p,final,authoryear]{elsarticle}

\usepackage[nopar]{lipsum}
\usepackage{multicol}

\usepackage{makecell}
\usepackage{multirow}
\usepackage{colortbl}
\usepackage[font=footnotesize]{subcaption}
\usepackage{booktabs}  
\usepackage[dvipsnames, table]{xcolor}
\usepackage[hyphens]{url}
\usepackage{hyperref}
\hypersetup{
  colorlinks,
  citecolor=blue,
  filecolor=blue,
  linkcolor=blue,
  urlcolor=blue
}
\usepackage{graphicx}
\usepackage{svg}
\usepackage[percent]{overpic}
\usepackage{pdfpages}
\graphicspath{{./Graphics/}}
\usepackage{amsfonts,amsmath,amsthm,amssymb,bm}



\usepackage{mathtools}

\newcommand\Item[1][]{%
  \ifx\relax#1\relax \item \else \item[#1] \fi
\abovedisplayskip=0pt\abovedisplayshortskip=0pt~\vspace*{-\baselineskip}}

\usepackage{enumitem}
\usepackage{footnote}
\usepackage{listings,verbatim}
\lstdefinestyle{custommatlab}{
  language=Matlab,%
  breaklines=true,%
  basicstyle=\footnotesize,
  morekeywords={matlab2tikz},
  keywordstyle=\color{blue},%
  morekeywords=[2]{1}, keywordstyle=[2]{\color{black}},
  identifierstyle=\color{black},%
  stringstyle=\color{matlilas},
  commentstyle=\color{matgreen},%
  showstringspaces=false,
  emph=[1]{for,end,break},emphstyle=[1]\color{red}, 
}
\lstdefinestyle{customphp}{
  language        = php,
  basicstyle      = \small\ttfamily,
  keywordstyle    = \color{dkblue},
  stringstyle     = \color{red},
  identifierstyle = \color{dkgreen},
  commentstyle    = \color{gray},
  emph            =[1]{php},
  emphstyle       =[1]\color{black},
  emph            =[2]{if,and,or,else},
  emphstyle       =[2]\color{dkyellow}}
\lstdefinestyle{customc}{
  breaklines=true, breakindent=20pt,
  frame=leftline,
  xleftmargin=\parindent,
  numbers=left,
  language=C, numberstyle=\tiny, numbersep=10pt,
  showstringspaces=false,
  basicstyle=\footnotesize\ttfamily,
  keywordstyle=\bfseries\color{green!40!black},
  commentstyle=\itshape\color{purple!40!black},
  identifierstyle=\color{blue},
  stringstyle=\color{orange},
  captionpos=t
}

\usepackage{fontawesome,marvosym}
\usepackage{tipa} 
\usepackage{fourier}    
\DeclareMathAlphabet{\mathcal}{OT1}{pzc}{m}{it}
\DeclareSymbolFont{letters}{OML}{cmm}{m}{it}
\definecolor{bgblue}  {rgb} {0.4, 0.4, 1.0}
\definecolor{bgblue}{rgb}{0.41961,0.90784,0.90784}%

\definecolor{bgred}   {rgb} {1.0, 0.4, 0.4}
\definecolor{bgred}{rgb}{1,0.61569,0.61569}%

\definecolor{bggreen} {rgb} {0.4, 1.0, 0.4}

\definecolor{fgblue}  {rgb} {0.0, 0.0, 0.4}
\definecolor{fgblue}{rgb}{0,0,0.6}%

\definecolor{fgred}   {rgb} {0.4, 0.0, 0.0}
\definecolor{fgred}{rgb}{0.6,0,0}%

\definecolor{fggreen} {rgb} {0.0, 0.4, 0.0}

\definecolor{dkgreen} {rgb} {0.0, 0.6, 0.0}
\definecolor{dkblue}  {rgb} {0.0, 0.0, 0.6}
\definecolor{dkyellow}{cmyk}{0.0, 0.0, 0.8, 0.3}
\definecolor{matgreen}{RGB} {28,  172,0}
\definecolor{matlilas}{RGB} {170, 55,241}
\definecolor{myBlue}{HTML}{7982db}
\definecolor{flatBlue}{HTML}{B2B2FF}
\definecolor{flatRed}{HTML}{FFB2B2}
\definecolor{flatBrown}{HTML}{EAD8C5}
\definecolor{flatGrey}{HTML}{E1E3E3}
\definecolor{new_blue}{HTML}{517EB9}
\definecolor{hl}{HTML}{FFEF36}

\DeclareSymbolFont{letters}{OML}{cmm}{m}{it}

\usepackage[nopostdot,nonumberlist,numberedsection,section=subsection]{glossaries}
\makenoidxglossaries

\newglossaryentry{breakpoint}
{
  name=breakpoint,
  description={the beginning or end point of a structural variation}
}

\newglossaryentry{copy number}
{
  name=copy number,
  description={number of copies of a particular gene}
}

\newglossaryentry{copy number alteration}
{
  name=copy number alteration,
  description={a change in copy number that has arisen in any cell of the body after conception}
}

\newglossaryentry{chromosome}
{
  name=chromosome,
  description={a molecule that contains part of the \gls{genome} in a condensed, manageable package}
}

\newglossaryentry{chromosomal inversion}
{
  name=chromosomal inversion,
  description={mutation in which a segment of a chromosome is reversed}
}

\newglossaryentry{fusion} { name=fusion, description={when two previously independent genes are
combined or fused together that results from \gls{chromosomal translocation}, \gls{interstitial
deletion}, and \gls{chromosomal inversion}} }

\newglossaryentry{genome}
{
  name=genome,
  description={genetic material of an organism}
}

\newglossaryentry{gene family} { name=gene family, description={a set of several similar genes
formed by duplication, generally with similar function} }

\newglossaryentry{nucleotide}
{
  name=nucleotide,
  description={form the basic structural unit of DNA. Encoded as either A,T,G, or C}
}

\newglossaryentry{HUGO}
{
  name=HUGO,
  description={HUGO(HUman Genome Organization) Gene Nomenclature Committee}
}

\newglossaryentry{HUGO symbol}
{
  name=HUGO symbol,
  description={unique gene identifier derived from the \href{https://www.genenames.org/about/guidelines/}{HUGO gene nomenclature guidelines}}
}

\newglossaryentry{in-frame} { name=in-frame, description={a mutation, where the translation into
protein is not completely disrupted, creating still-functional proteins} }

\newglossaryentry{intergenic}
{
  name=intergenic DNA,
  description={DNA located between genes and are noncoding (i.e. do not encode protein sequences), but occasionally control genes nearby}
}

\newglossaryentry{interstitial deletion} { name=interstitial deletion, description={a mutation in
which part of the DNA (not including the terminal portion of a chromosome) is left out during DNA
replication} }

\newglossaryentry{mutation}
{
  name=mutation,
  description={alteration of the nucleotide sequence of the genome}
}

\newglossaryentry{metastasis}
{
  name=metastasis,
  description={secondary malignant growth distant from the primary site of cancer}
}

\newglossaryentry{mutation signatures}
{
  name=mutation signatures,
  description={characteristic combinations of mutations arising from specific mutation processes}
}

\newglossaryentry{homologous chromosomes}
{
  name=homologous chromosomes,
  description={two chromosomes that carry the same genes, one from each parental source}
}

\newglossaryentry{non-homologous chromosomes}
{
  name=non-homologous chromosomes,
  description={two chromosomes that do not carry the same genes in contrast to \gls{homologous chromosomes}}
}

\newglossaryentry{out-of-frame} { name=out-of-frame, description={a mutation, where the translation
into a protein is completely disrupted, creating non-functional proteins.} }

\newglossaryentry{primary site}
{
  name=primary site,
  description={the location on the body where the first tumor progression begins}
}

\newglossaryentry{somatic}
{
  name=somatic mutation,
  description={genetic alteration acquired by cells that are the progeny of cancerous cells}
}

\newglossaryentry{structural variation}
{
  name=structural variation,
  description={any kind of structural variation to the genome}
}

\newglossaryentry{chromosomal translocation} { name=chromosomal translocation, description={two
types; Robertsonian translocation, which occurs when two \gls{non-homologous chromosomes} get
attached, and reciprocal translocation, which occurs when parts are exchanged between two non-homologous
chromosomes} }

\usepackage{pgf,pgfplots,pgfplotstable,tikz,tkz-fct}
\usepackage{tikz-timing}[2009/12/09]
\usepackage{float,fp}
\usepgflibrary{shapes}
\usepgfplotslibrary{external,statistics}
\usetikzlibrary{calc,trees,positioning,arrows,chains,shapes.geometric,decorations.pathreplacing,decorations.pathmorphing,shapes,matrix,shapes.symbols,er,backgrounds,decorations.text,spy,automata}

\pgfplotsset{compat=1.11}
\pgfplotsset{width=7cm,compat=1.13}

\def\getangle(#1) (#2)#3{%
  \begingroup%
  \pgftransformreset%
  \pgfmathanglebetweenpoints{\pgfpointanchor{#1}{center}}{\pgfpointanchor{#2}{center}}%
  \expandafter\xdef\csname angle#3\endcsname{\pgfmathresult}%
  \endgroup%
}

\tikzset{
  >=stealth',
  punktchain/.style={
    font=\tiny,
    rectangle,
    rounded corners,
    draw=black, very thick,
    text width=7em,
    minimum height=1em,
    text centered},
  line/.style={draw, thick, <-},
  element/.style={
    tape,
    top color=white,
    bottom color=blue!50!black!60!,
    minimum width=8em,
    draw=blue!40!black!90, very thick,
    text width=10em,
    minimum height=1em,
    text centered},
  every join/.style={->, thick,shorten >=1pt},
  decoration={brace},
  tuborg/.style={decorate},
  tubnode/.style={midway, right=2pt},
}

\begin{document}

\newcommand{\APPENDIX}{Appendix}

\begin{frontmatter} \title{Machine Learning in Precision Medicine to Preserve Privacy via
    Encryption}

  \author[UVIC]{William Briguglio}
  \ead{wbriguglio@uvic.ca}

  \author[UVIC]{Parisa Moghaddam}
  \ead{ParisaMoghaddam@uvic.ca}

  \author[UVIC,HU]{Waleed~A.~Yousef\corref{cor1}}
  \ead{wyousef@UVIC.ca, wyousef@fci.Helwan.edu.eg}
  \cortext[cor1]{Corresponding Author}

  \author[UVIC]{Issa Traor\'e}
  \ead{itraore@ece.uvic.ca}

  \author[NRC]{Mohammad Mamun}
  \ead{Mohammad.Mamun@nrc-cnrc.gc.ca}

  \address[UVIC]{ECE dep., University of Victoria, Victoria, BC, Canada.}
  \address[HU]{CS dep., Helwan University, Cairo, Egypt.}
  \address[NRC]{National Research Council of Canada, NB, Fredericton, Canada.}

  \begin{abstract}
    Precision medicine is an emerging approach for disease treatment and prevention that delivers
    personalized care to individual patients by considering their genetic makeups, medical
    histories, environments, and lifestyles. Despite the rapid advancement of precision medicine and
    its considerable promise, several underlying technological challenges remain unsolved. One such
    challenge of great importance is the security and privacy of precision health–related data, such
    as genomic data and electronic health records, which stifle collaboration and hamper the full
    potential of machine-learning (ML) algorithms. To preserve data privacy while providing ML
    solutions, this article makes three contributions. First, we propose a generic machine learning
    with encryption (MLE) framework, which we used to build an ML model that predicts cancer from
    one of the most recent comprehensive genomics datasets in the field. Second, our framework’s
    prediction accuracy is slightly higher than that of the most recent studies conducted on the
    same dataset, yet it maintains the privacy of the patients’ genomic data. Third, to facilitate
    the validation, reproduction, and extension of this work, we provide an open-source repository
    that contains the design and implementation of the framework, all the ML experiments and code,
    and the final predictive model deployed to a free cloud service.
  \end{abstract}


\end{frontmatter}


\section{Introduction}\label{sec:introduction}
Precision medicine is a departure from one-size-fits-all medicine toward the customization of
disease treatment and prevention for individuals by leveraging their individual variability in
genes, environment, and lifestyle. Two key aspects of the digital revolution of the last decade have
been an exponential increase in the amount of data being generated and a parallel increase in
hardware and GPU performance. While significant progress has been made in storing huge amounts of
data via big-data technologies, the ability to develop actionable knowledge that facilitates
individualized care through precision medicine has lagged behind.

Despite the exciting prospects of precision healthcare, it faces several technical and societal
hurdles related to the identification of health risks, diagnoses, and outcomes by analyzing data
extracted from integrated biomedical databases. Security and privacy concerns are such hurdles.
While precision health provides tremendous benefits by enabling better care, it can lead to personal
privacy breaches through genetic disclosure or genetic discrimination. To deliver targeted,
personalized care, personal data (e.g., specific human genome sequencing) must be shared with many
professionals in possibly diverse geographic locations or jurisdictions and sometimes over
unreliable channels, such as the internet. This poses several risks, such as insider threats, social
engineering, distributed denial of service (DDoS), illicit data inferences, cyber
bullying/blackmailing, etc.~\citep{Rahman2019PrivacyPreservingServiceSelection}. Unlike protected
health information (PHI), precision health data, such as genomic data, not only identifies patients
but also multiple generations of their families. Hence, such data can be leveraged to conduct
targeted security and privacy attacks against vulnerable individuals or groups of related
individuals if fallen in the hands of malicious actors.

In principle, a machine-learning (ML) model can be trained on either confidential or public data,
allowing more training samples, data distributions, and therefore more complex, predictive, and
generalizing models. These complex models can theoretically achieve higher predictive performance
and find novel associations within precision healthcare. However, performing analytics on new cases
provided by hospitals or medical centers should be treated with the utmost privacy preservation
level for the reasons introduced above.

From this perspective, the purpose of the present article is threefold. First, we propose a machine
learning with encryption (MLE) framework that considers these requirements and constraints and
facilitates preforming analysis on a real precision healthcare dataset while preserving its privacy.
Second, we illustrate the model's success by considering a case study using the MSK-IMPACT dataset,
one of the most recent comprehensive genomic datasets in the
field~\citep{Zehir2017MutationalLandscapeMetastaticCancer}, and we produce a predictive model for
cancer with higher accuracy than the most recent publications on the same
dataset\linebreak\citep{Penson2020DevelopmentGenomeDerivedTumor}; while preserving the privacy of
the cases as required. Third, to facilitate the capabilities of different communities, such as
software engineering, ML, and precision medicine, and for the extension and validation of this work,
we provide the following as an open-source repository: (1) the system design and implementation of
the MLE framework, (2) all the ML code and experiments on the MSK-IMPACT dataset, and (3) a free
cloud service for medical practitioners to predict their own cases using the MLE framework.

\bigskip

The remainder of the present article is as follows. Section~\ref{sec:related-work} introduces
homomorphic encryption (HE) and related work on the joint field of ML and HE.
Section~\ref{sec:fram-ml-encrypt} presents our proposed MLE framework and its main modules.
Section~\ref{sec:experiments} introduces the MSK-IMPACT dataset as a case study, describes the ML
experiments that we conducted on this dataset to provide a predictive model with privacy
preservation via HE, and introduces the open-source repository for this work.
Section~\ref{sec:conclusion} concludes the article and suggests future work that would complement
the current work. For the present article to be self-contained, the online supplementary material
provides more explanation of the basics of genomics necessary for our article as well as an example
of HE.


\section{Related Work and Literature Review}\label{sec:related-work}
Any privacy-preserving precision health analytics builds on two main components: (1) the security
and privacy features required to protect interactions with the data by stakeholders and (2) ML
predictive models for this data. Each is reviewed briefly below.

\subsection{Homomorphic Encryption}\label{sec:homom-encrypt}
Encryption is the process of converting data from something intelligible into something
unintelligible by sealing data in a metaphorical vault that can only be opened by somebody holding
the secret decryption key to prevent unauthorized personnel from viewing them. A special scheme of
encryption is HE, which was originally proposed by
\cite{Rivest1978OnDataBanksAndPrivacyHomomorphism} as a way to encrypt data such that certain
operations could be performed on them without possessing that secret key (i.e., without decryption).
The term ``homomorphic encryption'' describes a class of encryption algorithms that satisfies the
homomorphic property; that is, certain operations, such as addition, can be carried out on
ciphertext directly so that upon decryption the same answer is obtained as operating on the original
messages. Therefore, HE allows other parties (e.g., the cloud and service providers) to calculate
certain mathematical functions expressed only in terms of these operations on the encrypted data
while preserving the function and format of the encrypted data. For brevity, these types of
functions are referred to as ``HE-friendly''. Formally, this can be expressed as
\begin{align}
  Dec[ k_{s},\ Enc(k_{p}, m_{1}) \diamond Enc(k_{p}, m_{2})] = m_{1} \circ m_{2},\label{eq:FHE}
\end{align}
where $k_{s},\ k_{p}$ are the secret and public keys, respectively (since they are not equal, this is
called ``asymmetric encryption''), $m_{1}, m_{2} \in M$ are two values on which we wish to perform
encrypted operations on, $M$ is the message space of the HE\linebreak scheme (i.e., the set of all
possible values acceptable by the scheme), and $\diamond, \circ$ are operations in encrypted and
plain-text space, respectively. Like other types of encryption schemes, HE has three main functions:
key generation, message encoding, and decoding (the supplementary material provides some details on
these operations.)

The remarkable properties of HE schemes are not without limitations. First, the set of functions
that can be computed in ciphertext space is very restricted. Second, the computational complexity of
the HE scheme depends primarily on the level of multiplication (i.e., the degree of the polynomial
being evaluated) carried out on the encrypted data. Third, the ciphertext size increases
considerably after encryption. Fourth, random noise is added to encrypted values for security
reasons that varies with the type of operation (e.g., multiplication increases noise much more than
addition). If this noise grows too large, then decryption yields incorrect results. There are three
basic approaches to implement HE~\citep[see,
e.g.,][]{Acar2018AsurveyOnHomomorphicEncryptionSchemes}:

\begin{description}[partopsep=0.1in,parsep=0in,topsep=0in,itemsep=0.1in,leftmargin=0in]
  \item[partially homomorphic encryption (PHE),] which allows only one type of operation to be
  executed on encrypted values an unlimited number of times.

  \item[somewhat homomorphic encryption (SWHE),] where the size of the ciphertext grows with each
  homomorphic operation and hence the maximum number of allowed homomorphic operations is limited.

  \item[fully homomorphic encryption (FHE),] which supports an unlimited number of additions and
  multiplications~\citep{Gentry2009FullyHomomorphicEncryptionScheme}. This property makes FHE the
  most sophisticated HE scheme and the ``holy grail'' of modern cryptography. The FHE scheme
  supports basic arithmetic computations on encrypted data. Despite being a potential cryptographic
  technique, however, some FHE schemes remain impractical for real-world applications due to their
  computational overhead.
\end{description}

\subsection{Machine Learning with Encryption}\label{sec:encrypt-mach-learn}
\begin{table}[t]\centering
  \caption{The eight possible scenarios of encrypting the three components: the training dataset
    $\mathbf{tr}$, the ML model parameters $\mathcal{M}$, and the testing dataset
    $\mathbf{ts}$.}\label{tab:eight-poss-comb}
  \resizebox{\columnwidth}{!}{%
    \begin{tabular}{l@{}cllll}
      \toprule
      \#\hspace{5pt}  &$\mathbf{tr}\ \mathcal{M}\ \mathbf{ts}$ & Literature &Dataset & ML & Enc. Library\\
      \midrule
      0& 0 0 0 & Ordinary ML&&&\\[5pt]
      \multirow[t]{3}{*}{1}& \multirow[t]{3}{*}{0 0 1} &  Our present article&MSK &many& SEAL\\
                      &&\cite{Dowlin2016CryptoNetsApplyingNeuralNetworksToEncryptedData} &MINIST &NN &SEAL\\
                      &&\cite{Hesamifard2017DNNOverEncryptedData} &MINIST, CIFAR-10 &DNN &Helib\\[5pt]
      2& 0 1 0 & \cite{Bost2015MachineLearningClassificationOverEncryptedData}& Multiple & NB, HP, DT & self-implementation\\[5pt]
      3& 0 1 1 &  \multicolumn{4}{l}{------}\\[5pt]
      \multirow[t]{2}{*}{4}& \multirow[t]{3}{*}{1 0 0} &\cite{Graepel2013MlConfidentialMachineLearningOnEncryptedData} & Wisconsin& FDA& Magma\\
                      && \cite{Aslett2015EncryptedStatisticalMachineLearning}& Multiple& NB, RF& EncryptedStats\\
                      &&\cite{Nandakumar2019TowardsDNNTrainingOnEncryptedData}& MNIST6 &DNN7 &HElib\\[5pt]
      5& 1 0 1 & \multicolumn{4}{l}{------ Not possible under the current theory}\\[5pt]
      6& 1 1 0 & \multicolumn{4}{l}{\multirow{2}{*}{------ Not practical: training on encrypted data already produces encrypted model}}\\
      7& 1 1 1 & &&&\\
      \bottomrule
    \end{tabular}}
\end{table}
Any ML algorithm trains on some training dataset $\mathbf{tr}$, fits a model`s parameters
$\mathcal{M}$, and finally tests on a testing dataset $\mathbf{ts}$. Therefore, in principle, there
are eight possible combinations or scenarios to introduce privacy via encryption to the learning
process by encrypting (or leaving unencrypted) each of these three components.
Table~\ref{tab:eight-poss-comb} summarizes those eight scenarios; below, we provide more details on
them and their connections to the present article. We use 0 to denote an unencrypted component,
where it still can only be encrypted using the public key $k_p$ of another component without having
access to its private key $k_s$, and we use 1 to denote an encrypted component, where its private
key $k_s$ is not available for the other two components.

Scenario 0 is denoted by the binary combination 000, when $\mathbf{tr}$, $\mathcal{M}$, and
$\mathbf{ts}$ are all not encrypted; this is the typical ML paradigm, where no privacy is of
concern.

Scenario 1 is denoted by 001, where only $\mathbf{ts}$ needs to be encrypted; this is the scenario
of the present article, as Sections~\ref{sec:fram-ml-encrypt} and \ref{sec:experiments} describe. In
such a scenario, since the model has access to an unencrypted training set, such as a public
dataset, only the client data $\mathbf{ts}$, which could be patients' genome records, are sensitive.
Although the standard homomorphic property as defined in ~\eqref{eq:FHE} would imply that
$\mathcal{M}$ must be encrypted with $k_p$ to predict on an encrypted $\mathbf{ts}$, this is not the
case for our work, which leverages special techniques implemented in SEAL
\cite{Microsoft2019Github}, that allow encryption with plain-text multiplication with the caveat
that the results themselves are encrypted and can only be decrypted with $k_s$. Research exists in
this category but in areas of application other than precision medicine.
\cite{Dowlin2016CryptoNetsApplyingNeuralNetworksToEncryptedData} showed that a cloud service is
capable of applying a neural network (NN) to encrypted $\mathbf{ts}$ to make encrypted predictions
and return them in encrypted forms. They constructed a convolution NN (CNN) model from the
unencrypted MINIST dataset and then produced a simpler FHE-friendly version of the CNN constructed
only from addition and multiplication operations so that the parameters could be encrypted using the
public key of the private testing dataset $\mathbf{ts}$. \cite{Hesamifard2017DNNOverEncryptedData}
developed new techniques to allow testing CNN on encrypted $\mathbf{ts}$. First, they designed
methods to approximate the activation functions commonly used in CNNs with low-degree, FHE-friendly
polynomials. Then, they trained a CNN on unencrypted $\mathbf{tr}$ with the approximation
polynomials instead of the original activation functions. Finally, they converted the trained CNN to
make predictions on encrypted $\mathbf{ts}$.

Scenario 2 is denoted by 010, where the model is trained on an unencrypted training dataset
$\mathbf{tr}$. However, the model parameters themselves are then encrypted, which may imply privacy
in $\mathbf{tr}$, as well if the training is pursued locally where $\mathbf{tr}$ resides. Although
the testing data $\mathbf{ts}$ is denoted by 0, it must be sent to $\mathcal{M}$ encrypted with its
public key, as it is not possible, according to the theory of FHE, to pursue binary operations on
encrypted numbers (parameters of $\mathcal{M}$) and unencrypted numbers ($\mathbf{ts}$), without the
results being encrypted and only decryptable with the same $k_s$ that can decrypt $\mathcal{M}$. The
virtue of scenario 2 is that it entails more freedom in choosing the model $\mathcal{M}$ as opposed
to scenarios 4--7, where $\mathbf{tr}$ is encrypted and a stringent limitation is incurred for
choosing the model $\mathcal{M}$ that can train on encrypted data. We are not aware of any
literature that applies scenario 2 explicitly; however,
\cite{Bost2015MachineLearningClassificationOverEncryptedData} provided a very nested, layered model
that could be classified as 010 scenario, but without relying solely on HE. They implemented a
decision tree, naive Bayes and hyperplane decision that could test (not train) on encrypted data and
built their models using cryptographic ``building blocks'' that emphasized protecting the model
parameters and test data. They also used garbled circuits to compare encrypted data, which allowed a
construction of argmax with alterations to ensure the ordering was not leaked. However, this
introduced $k-1$ round trips between the party performing argmax and the party that can perform
decryption. These building blocks allowed the implementation of decision tree, naive Bayes, and
hyperplane decision with some minor changes. The building blocks also allowed the construction of
other ML methods and a combination of methods using AdaBoost, which the authors demonstrated.

Scenario 3 is like 2 in that the model is trained on an unencrypted $\mathbf{tr}$, and
$\mathcal{M}$`s parameters are then encrypted; however, the test dataset $\mathbf{ts}$ is also
encrypted with a different $k_s$ than $\mathcal{M}$. Since there is no known method in the
literature that allows the use of binary operations on two numbers encrypted with different $k_s$,
scenario 3 (011) is not theoretically feasible under the current theory of cryptography.

Scenario 4 is denoted by 100, where an ML model is trained on encrypted $\mathbf{tr}$ (as in
scenarios 5--7, as well). Hence, the model $\mathcal{M}$ will have encrypted weights by product, and
the testing data must be sent to $\mathcal{M}$ encrypted with the same public key of $\mathbf{tr}$,
as explained above. Therefore, the reason scenario 5 (101) is not theoretically possible is the same
as scenario 3. Furthermore, scenarios 6 and 7 (11x) are not of any practical interest, since the
produced encrypted $\mathcal{M}$ does not need further encryption; this is possibly the reason for
the absence of literature on these two scenarios. Under scenario 4, \cite{Graepel2013Confidential}
defined a fully confidential version of linear means and Fisher’s discriminant analysis (FDA), which
can train and test on encrypted data. Linear means are rewritten to avoid division when learning the
weights. The resulting decision function returns a multiple of the original decision function with
the same sign. However, FDA requires the inverse of the covariance matrix to obtain the feature
weights. This is found using gradient descent, the $r^{th}$ iteration of which is shown to be a
$d$-degree polynomial, where $d = 2(r -1)+1$. \cite{Aslett2015EncryptedStatisticalMachineLearning}
provided a completely random forest (CRF) implementation that could train and test on encrypted
data. Among other alterations to the algorithm, the key difference was encoding feature values using
one hot encoding after quantile partitioning. CRFs have important benefits, especially learning
incrementally. The authors also provided a naive Bayes classifier that could train and test on
encrypted data. It avoided parametric Gaussian modeling of predictors by directly modeling the
decision boundary $x_{j}m_{j}+b_{j}$ for each predictor $X_{j}$. This required a homomorphic
implementation of regression, which they also provided. \cite{Hesamifard2017DNNOverEncryptedData}
discussed the computational complexity that makes NN training on encrypted data impractical despite
being theoretically possible. Other authors have targeted simpler ML algorithms to avoid heavy
computations of encrypted NNs. However,
\cite{Dowlin2016CryptoNetsApplyingNeuralNetworksToEncryptedData} show that training CNNs on
encrypted data is possible. If all the activation functions and the loss function are polynomials,
back-propagation can be computed using only addition and multiplication. However, high-degree
polynomials used during back-propagation make it computationally prohibitive.
\cite{Nandakumar2019TowardsDNNTrainingOnEncryptedData} evaluated the feasibility of training NNs on
encrypted data completely non-interactively. His proposed system used the FHE toolkit HElib to
implement stochastic gradient descent (SGD) for training. He used ``ciphertext packing'' to minimize
the number of required bootstrapping operations and to enable the parallelization of computations at
each neuron, thereby significantly reducing the computational complexity. This, in combination with
simplifying the network architecture, allowed him to practically train neural networks over
encrypted data despite the computational hurdles.


\section{A Machine Learning with Encryption (MLE) Framework}\label{sec:fram-ml-encrypt}
\tikzset{
  state/.style={
    rectangle,
    rounded corners,
    draw=black, very thick,
    minimum height=2em,
    inner sep=0.05in,
    text width = 0.5\columnwidth,
    font = \tiny
  },
}

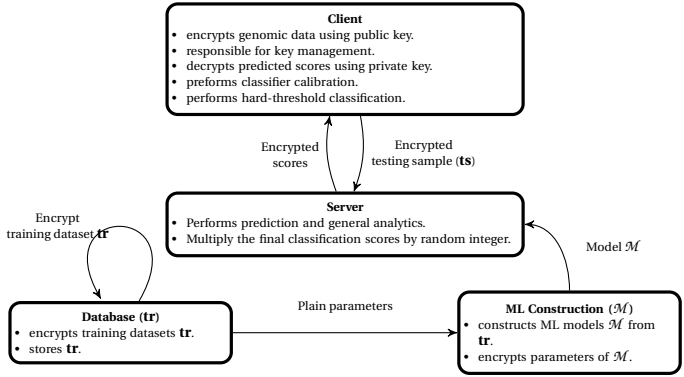
\begin{figure}[t]\centering
    \begin{tikzpicture}[->,>=stealth']
      \node[state](CLIENT)
      {\hfil\textbf{Client}\hfil\\
        \begin{itemize}[partopsep=0in,parsep=0in,topsep=0in,itemsep=0.0in,leftmargin=.05in,labelsep=0.05in]
          \item encrypts genomic data using public key.
          \item responsible for key management.
          \item decrypts predicted scores using private key.
          \item preforms classifier calibration.
          \item performs hard-threshold classification.
        \end{itemize}
      };

      \node[state, below=1cm of CLIENT.south, anchor=north](SERVER)
      {\hfil\textbf{Server}\hfil\\
        \begin{itemize}[partopsep=0in,parsep=0in,topsep=0in,itemsep=0.0in,leftmargin=.05in,labelsep=0.05in]
          \item Performs prediction and general analytics.
          \item Multiply the final classification scores by random integer.
        \end{itemize}
      };

      \node[state, below right=1cm and 0.5\columnwidth of SERVER.south, anchor=east, text width = 0.3\columnwidth](ML)
      {\hfil\textbf{ML Construction ($\mathcal{M}$)}\hfil\\
        \begin{itemize}[partopsep=0in,parsep=0in,topsep=0in,itemsep=0.0in,leftmargin=.05in,labelsep=0.05in]
          \item constructs ML models $\mathcal{M}$ from $\mathbf{tr}$.
          \item encrypts parameters of $\mathcal{M}$.
        \end{itemize}};

      \node[state, below left=1cm and 0.5\columnwidth of SERVER.south, anchor=west, text width = 0.3\columnwidth](DB)
      {\hfil\textbf{Database ($\mathbf{tr}$)}\\
        \begin{itemize}[partopsep=0in,parsep=0in,topsep=0in,itemsep=0.0in,leftmargin=.05in,labelsep=0.05in]
          \item encrypts training datasets $\mathbf{tr}$.
          \item stores $\mathbf{tr}$.
        \end{itemize}};

      \path[->]
      (CLIENT) 	edge[bend left=15, align=center, right = 0.5cm]  node[font = \tiny]{Encrypted\\ testing sample ($\mathbf{ts}$)} (SERVER)
      (SERVER) 	edge[bend left=15, align=center, left = 0.5cm]  node[font = \tiny]{Encrypted\\ scores} (CLIENT)
      (ML)  	edge[out=north, in=east, align=center] node[right=0.25cm,font = \tiny]{Model $\mathcal{M}$} (SERVER)
      (DB)  	edge[] node[above = 0.15cm,font = \tiny]{Plain parameters} (ML)
      (DB)  	edge[in=120, out=60, loop, align=center] node[left = 0.05cm, font = \tiny]{Encrypt \\ training dataset $\mathbf{tr}$} (DB);
    \end{tikzpicture}
    \caption{A block diagram of a privacy-preserving MLE framework for precision medicine that can
      accommodate any of the eight scenarios of Table~\ref{tab:eight-poss-comb}. Each block contains
      the functionalities that can be performed in that block depending on the adopted
      scenario.}\label{fig:BD}
\end{figure}


In this section we propose a simple four-component system architecture to accommodate any of the
eight scenarios of encryption in Table~\ref{tab:eight-poss-comb}. This simple
architecture, illustrated in Figure~\ref{fig:BD}, fulfills the privacy-preserving requirements that
are mandatory for future ML-based precision medicine. The components of this architecture are
explained below.

\begin{description}[partopsep=0in,parsep=0in,topsep=0.1in,itemsep=0.1in,leftmargin=0in]
  \item[\textit{Database} ($\mathbf{tr}$)] is a reservoir for both publicly available genetic
  datasets, which do not require preserving privacy, and private datasets, which need encryption
  prior to public sharing. Whenever a new dataset is revealed, it can be added to this reservoir for
  more accurate future analytics.

  \item[\textit{ML Construction} ($\mathcal{M}$)] is the engine that constructs models---including
  transformation, feature selection, resampling, etc.---from the datasets in the \textit{database}
  module. This module can be open-sourced for the entire community and can always be updated as new
  ML methods merge or more accurate models are constructed. In addition, the module can train on its
  own private dataset, which is not part of the \textit{database} module, and then encrypt its model
  parameters $\mathcal{M}$. Alternatively, it can establish a protocol with the \textit{database}
  module to train on the private dataset without encryption for a wider range of algorithms and then
  encrypt the model parameters to preserve the dataset's privacy (Cases 01x in
  Table~\ref{tab:eight-poss-comb}).

  \item[\textit{Client} ($\mathbf{ts}$)] is where the testing data, which is probably sensitive and
  confidential, resides and needs analytics. The owner of this data can opt to encrypt it, and this
  encryption can be provided via simple software components installed on the \textit{client} side
  available via communication with the \textit{server}. Next, the encrypted testing data is sent to
  the \textit{server} for prediction. Finally, the encrypted predicted scores are received back. The
  \textit{client} should be responsible for setting the threshold on the scores for the final hard
  decision or classification. This is to achieve a required level of aggressiveness to control the
  per-class sensitivity, such as in the case of the binary classification problem, in which the
  threshold provides the trade-off between the sensitivity and specificity and thus controls the
  operating point on the receiver operating characteristic (ROC) curve.

  \item[\textit{Server}] is the cloud engine for prediction. On the one hand, it interfaces with the
  \textit{client} to receive the encrypted dataset for prediction and sends back encrypted
  predictions, and on the other hand, it interfaces with \textit{ML construction} to receive a
  particular predictive model. Based on the underlying encryption scenario
  (Table~\ref{tab:eight-poss-comb}), the \textit{server} receives the appropriate public key from
  these two modules. In addition, for the $C$-class classification problem and for greater privacy
  preservation for the model and/or the dataset ($\mathbf{tr}$), the server can optionally multiply
  the scores $s_c(x), c = 1, \ldots,C$, where $x \in \mathbf{ts}$, by a random integer. This keeps
  the relative $C$ scores unaffected. However, this disallows the \textit{client} from inferring
  information about the model weights ($\mathcal{M}$) by sending pseudo-feature vectors in the form
  $x = (0, \ldots,1,0,\ldots)$ (only one feature is 1; the others are zeros).
\end{description}

To illustrate the utility of this simple architecture, we demonstrate how scenario 1 can be
implemented in a very practical setup. When ML training is based on public data ($\mathbf{tr}$), the
weights of the trained model $\mathcal{M}$ are deployed on the \textit{server} in unencrypted form,
while the queries ($\mathbf{ts}$) must be encrypted for security sensitivity. Under a hospital’s
public key, many parties may also be eligible to upload data (e.g., doctors and patients). The
\textit{server} is used for deploying ML implementations $\mathcal{M}$. In this case, the hospital
sends encrypted data to the \textit{server}. In the \textit{server}, many computations can be done
on the encrypted data and the results sent back to the hospital. Only the hospital can decrypt the
data because the private key is provided only on the hospital side. In the following section, we
conduct a large set of ML experiments under the MLE framework and scenario 1 on one of the most
recent high-quality genomic datasets.


\section{Experiments}\label{sec:experiments}
\newcolumntype{L}[1]{>{\raggedright\arraybackslash}m{#1}}
\newcolumntype{C}[1]{>{\centering\arraybackslash}m{#1}}
\newcolumntype{R}[1]{>{\raggedleft\arraybackslash}m{#1}}
\begin{table*}[t]\centering
  \caption{Different configurations of feature pre-processing and classifiers tried on the dataset:
    $5\times 2 \times 4 \times (8 + 36 + 12) = 2,240$ different
    configurations.}\label{tab:diff-conf-feat}
  \resizebox{\textwidth}{!}{%
  \begin{tabular}{cccp{.01in}cL{3.3in}}
    \toprule
    \multicolumn{3}{c}{Feature Preprocessing} &&\multirow{2}{*}{Classifier}                           & \multicolumn{1}{c}{\multirow{2}{*}{Parameters}}\\
    \cline{1-3}
    Transformation &Selection & Reduced $p$ &&&\\
    \midrule
    \multirowcell{5}{None\\ Standardize\\ MinMax\\ log($x$)\\ $\log(x+1)$} & \multirowcell{5}{MI\\ $\chi^2$} & \multirowcell{5}{0\\ 2500\\ 3500\\ 4500} && LR & penalty: l1, l2;\quad C: 0.1, 1, 10;\quad solver: liblinear, lbfgs\\
    \cline{5-6}
                                              &&&& SVM & kernel: linear, poly, rbf;\quad penalty: l1, l2;\quad C: 0.1, 1, 10;\quad loss: hinge, squared\_hinge\\
    \cline{5-6}
                                              &&&&RF & criterion: gini, entropy;\quad n\_estimators: 100, 200, 300;\quad bootstrap: False, True\\
    \bottomrule
  \end{tabular}}
\end{table*}

\begin{table*}[h]\centering
  \caption{The best three configurations in Table~\ref{tab:diff-conf-feat}. The first two are the LR
    and SVM, which are HE-friendly.}\label{tab:winn-conf}
  \resizebox{\textwidth}{!}{%
  \begin{tabular}{cccp{.01in}cL{3.2in}c}
    \toprule
    \multicolumn{3}{c}{Feature Preprocessing} &                   & \multirow{2}{*}{Classifier} & \multicolumn{1}{c}{\multirow{2}{*}{Parameters}} & \multirowcell{2}{Accuracy\\(\%)}                                                                                \\
    \cline{1-3}
    Transformation                            & Selection         & Reduced $p$                 &                                                 &     &                                                                            &       \\
    \midrule
    Standardize                               & $\chi^2$           & 2500                        &                                                 & LR  & C: 1; Penalty: l2; Solver: liblinear                           & 77.47 \\
    None                                      & $\chi^2$           & 2500                        &                                                 & SVM & kernel: linear; C:0.1; penalty: l2; loss: squared\_hinge & 77.23 \\
    Standardize                               & MI & 2500                        &                                                 & RF  & criterion: gini; n\_estimators: 200; bootstrap: False          & 73.92 \\
    \bottomrule
  \end{tabular}}
\end{table*}

\begin{table}[h]\small\centering
  \caption{Accuracy and percentage of score ranking of the best model at different multiplier
    precision.}\label{tab:accur-perc-score}
  \resizebox{\columnwidth}{!}{%
  \begin{tabular}{rccccccccc}
    \toprule
    Precision& $10^1$& $10^2$& $10^3$& $10^4$& $10^5$& $10^6$& $10^7$& $10^8$& $10^9$\\
    Accuracy (\%)& 73.88& 77.41& 77.47& 77.41& 77.41& 77.41& 77.41& 77.41& 77.41\\
    agreeing ranks (\%)& 36.49& 70.80& 95.82& 99.41& 99.94& 100.0&  100.0&  100.0&  100.0\\
    \bottomrule
  \end{tabular}}
\end{table}


In this section, we concisely describe the dataset (more details are provided in the supplementary
materials), explain the different ML experiments to build the predictive model, demonstrate
computational aspects of the encryption process, and finally introduce the open-source platform of
the whole project.

\subsection{MSK-IMPACT Dataset}\label{sec:dataset}
MSK-IMPACT, a clinical sequencing cohort
dataset~\citep{Zehir2017MutationalLandscapeMetastaticCancer}, comprises genomic patient records
extracted from tumor-tissue samples taken from 10,336 patients. Since tumors are usually the results
of many mutations, there are more than 100,000 discovered mutations. The dataset consists of 11
files linked together with \texttt{sample\_ID} and \texttt{Patient\_ID} and contains various
information about the \gls{somatic}\textcolor{blue}{s} mutations within the genomic samples,
including mutation signature, copy number alternation, and gene fusion data files. ``\textit{With
  maturing clinical annotation of treatment response and disease-specific outcome'', according
  to~\citep{Zehir2017MutationalLandscapeMetastaticCancer}, ``this dataset will prove a
  transformative resource for identifying novel biomarkers to inform prognosis and predict response
  and resistance to therapy...Tumor molecular profiling is a fundamental component of precision
  oncology, enabling the identification of genomic alterations in genes and pathways that can be
  targeted therapeutically}''.

The authors of the dataset tried to associate ``biomarkers'' with a particular type of cancer using
simple methods of association, such as relative frequency. Then, to illustrate the usefulness of
their DNA-sequence approach, they leveraged the Oncology Knowledge
Base~\citep{Chakravarty2017Oncokb} to see how many of the mutations they detected (stratified by
cancer type) were known to be actionable, that is, have an associated treatment or gene therapy.
Recently, a subset of the MSK-IMPACT-2017 dataset with a portion of these features, somewhat lightly
engineered, was used by~\cite{Penson2020DevelopmentGenomeDerivedTumor} with a more ML-oriented
approach. Using logistic regression (LR), they achieved an overall accuracy of 75\% in detecting the
cancer type from genetic information; yet, their approach did not consider privacy preservation.

\subsection{Building the Model}\label{sec:building-model}
In addition to the predictive power required for any ML model, the objective of privacy preservation
via FHE requires the final ML model be FHE-friendly, that is, based only on addition and
multiplication operations, as was explained in Section~\ref{sec:related-work}. Some ML models cannot
satisfy both of these objectives. For example, a random forest (RF) has binary decision splits that
are not FHE-friendly. However, although linear models (LM), logistic regression (LR), support vector
machines (SVM), and many others are all HE-friendly, they may not perform well on a particular
dataset.

For our MSK dataset, we tried 2,240 different ML configurations that were the cross products of five
methods for transforming features, two methods for dropping the least important features, four
values for the number of dropped features $p$, and three classifiers, LR, SVM, RF, each with several
sets of tuning parameters. The detailed parameters of these 2,240 experiments are listed in
Table~\ref{tab:diff-conf-feat}. The best three configurations of the 2,240 are listed in
Table~\ref{tab:winn-conf} and were achieved by LR, SVM, and RF, respectively. LR, after feature
standardization, $\chi^2$ selection, and dropping the least informative 2,500 features, achieved the
highest accuracy of 77.47\% (using 10-fold cross validation), which was higher than that obtained
by~\cite{Penson2020DevelopmentGenomeDerivedTumor} on the same dataset.

Although the model was trained on MSK, a public dataset that requires no encryption, the final model
parameters (Eq.~\eqref{eq:LRweights}) must be encrypted since they will be multiplied by the
encrypted features of the testing data. The next section explains the encryption of both the testing
dataset and the parameters of the final trained model. Since we are using FHE, we must convert all
floats to integers. To do this, we scale all floats by some $10^d$, where $d$ is the number of
decimal places included in the scaled floats, therefore controlling the computational precision.
After scaling, we round off any remaining decimal to achieve the final integers. The effect of
precision on the accuracy of the best model is illustrated in Table~\ref{tab:accur-perc-score},
where it is clear that a multiplier of $10^4$ would be adequate.

\subsection{Encrypting the Testing Dataset}\label{sec:encrypt-test-datas}
\subsubsection{The SEAL library}\label{sec:seal-library}
Different libraries exist for implementing HE~\citep{Carey2020ExplanationImplementationThreeOpen};
among them, SEAL~\citep{Microsoft2019Github} is an open-source HE library developed by the
cryptography and privacy research group at Microsoft. The library is written in C++ and can run in
many environments. SEAL allows addition and multiplication to be performed on numbers. Other
operations, such as encrypted comparison, sorting, and regular expressions, are in most cases not
feasible on encrypted data using this technology. SEAL supports two FHE schemes: the
Brakerski/Fan-Vercauteren (BFV) scheme, which allows modular arithmetic to be performed on encrypted
integers, and the Cheon-Kim-Kim-Song (CKKS) scheme, which allows addition and multiplication on
encrypted real or complex numbers, but this latter scheme yields only approximate results.

\subsubsection{Computational Aspects of Encryption Operations}\label{sec:encrypt-oper}
\begin{table}[t]\centering
  \caption{Effect of encryption parameters on encryption time. All libraries support automatic
    selection of \texttt{CoeffModulus}. NA indicates noise $\sim 0$}\label{tab:diff-exper-with}
  \begin{tabular}{cccc}
    \toprule
    \# &polyModulusDegree &PlainModulus &Time in Sec.\\
    \midrule
    1 & 8192 & 2048 & 3456\\
    2 & 2048 & 1024 & 96\\
    3 & 1024 & 512 &  NA\\
    4 & 1024 & 1024 & NA\\
    5 & 2048 & 1024 & 87\\
    6 & 2048 & 512 & 88\\
    7 & 2048 & 1499 & 94\\
    8 & 2048 & 786433 & 89\\
    \bottomrule
  \end{tabular}
\end{table}
From the previous section, a $C$-class LR model was the winner for this dataset; formally, this model is given by
\begin{subequations}\label{eq:LRweights}
  \begin{align}
    \Pr(G = c | X = x) &= \frac{\exp{(s_c)}}{1+\sum_{l=1}^{C-1}\exp{(s_l)}},\label{eq:LRweights-a}\\
                         s_c &= w_{c0} + w_c' x,\ c = 1,\ldots,C,\label{eq:LRweights-b}
  \end{align}
\end{subequations}
where, $C=22$ types of cancer, the patient feature vector is $x \in \mathbf{R}^{p}, p=5,599$; and the
testing dataset $\mathbf{ts}$ to be encrypted has $N = 7,791$ patient records. By construction, the
MLE framework requires sending the encrypted score of each testing observation to the client rather
than the final hard decision for trading off the types of error. In addition,
from~\eqref{eq:LRweights}, the numerator is a monotonic exponential function and the denominator is
only for scaling, so probabilities sum to 1. Therefore, it is sufficient to encrypt the linear term
$s_c$ and treat it as the final score sent to the \textit{client}. We applied the BFV scheme
implementation of SEAL to perform this weighted summation term. The encryption operations are
explained as follows. (1) Scale and encrypt the feature list. (2) Multiply encrypted features by
plain text weights and sum encrypted values along with the scaled bias step. (3) Decrypt the results
and repeat step 2 for each set of coefficients, i.e. each class. We tested different encryption
parameters to compare computational time. Table~\ref{tab:diff-exper-with} illustrates the
computational time required as a function of a subset of the parameter space. Rows 3 and 4, caused
the ciphertext noise budget to reach zero. This noise budget is determined by the encryption
parameters, and once the noise budget of a ciphertext reaches zero, it becomes too corrupted to be
decrypted. Thus, it is essential to choose parameters large enough to support the desired
computations; otherwise, the correct result is impossible to obtain, even with the secret key. The
values in row five give the best average per sample prediction time after testing on the entire
dataset (7791 records), which spanned over seven days of computations on an i7core--2.5GHz--16G
machine. From Eq.~\eqref{eq:LRweights-b}, this time is obviously
$T = N C \bigl( (p+1)(E + M + A) + D\bigr)$, where $E$, $M$, $A$, and $D$ are the encryption,
multiplication, addition, and decryption times, respectively. During this experiment,
\texttt{IntegerEncoder} was used to encode integers to BFV plain-text polynomials.
\texttt{IntegerEncoder} is easy to understand and uses simple computations; however, there are more
efficient approaches such as BatchEncoder, which can be investigated in future works.

\subsection{MLE Framework: Opensource and Deployment}\label{sec:mle-fram-opens}
The official repository of this project~\citep{Briguglio2021MachineLearningViaEncryption} contains
three sets of open-source resources. (1) The Python code that produced all the ML experiments from
Section~\ref{sec:experiments} is offered, organized, and commented on to challenge the ML community
to develop more accurate, predictive models. (2) The system design and client-server implementation
and implementation code of the MLE framework from Section~\ref{sec:fram-ml-encrypt} is offered to
the software engineering community to propose more system functionalities. (3) A free cloud service
that encapsulates the best ML model and the client-server design is offered as a simple end-user
interface to individuals in the medical field to test on particular cases. We hope this kind of
dissemination to different communities helps the evolution of privacy-preserving ML for precision
medicine.


\section{Conclusion and Suggested Future Work}\label{sec:conclusion}
Toward building privacy preserving Machine Learning (ML) models for precision medicine, this article
has three contributions. First, we proposed and implemented a machine learning with encryption (MLE)
framework that accommodates different scenarios for encrypting the ML training-testing process.
Second, and most importantly, we analyzed the recent high-quality clinical sequencing cohort dataset
MSK-IMPACT and provided a predictive model that is both secure and outperforming the most recent
predictive model built for the same dataset. Third, we offered the ML, software engineering, and
precision medicine communities free resources: respectively, the client-server implementation of the
framework, the Python code of all the ML experiments, and a cloud service to test genomic cases.
These offerings contribute to the evolution of the privacy-preserving analytics of precision
medicine.

From Table~\ref{tab:eight-poss-comb}, it seems that scenario 2 is an open venue for incorporating
more accurate ML models on public datasets. As discussed in Section~\ref{sec:related-work},
\cite{Bost2015MachineLearningClassificationOverEncryptedData} implemented a very sophisticated
scenario that can be classified indirectly as scenario 2. They assumed that the owner of the dataset
$\mathbf{ts}$ is responsible for testing the model $\mathcal{M}$. They designed their model
encryption in a nested way to preserve its privacy and protect it from the owner of $\mathbf{ts}$ to
learn anything about its structure. Because of our MLE framework, we do not need these constraints,
since the computations run on the \textit{server} side, not the \textit{client} side. Therefore,
under the MLE framework, future work can benefit from scenario 2 by designing more ML models on
datasets and encrypting the model parameters if needed.


\section{Acknowledgment}\label{sec:acknoledgment}
The authors thank the National Research Council (NRC) for funding the project under the
Collaborative Research and Development Grant (award number CDTS-102).


\bibliographystyle{model2-names}
\bibliography{booksIhave,publications,../Bibliography/Bibliography/AA-Bibliography-AA.bib}

\begin{thebibliography}{19}
\expandafter\ifx\csname natexlab\endcsname\relax\def\natexlab#1{#1}\fi
\providecommand{\url}[1]{\texttt{#1}}
\providecommand{\href}[2]{#2}
\providecommand{\path}[1]{#1}
\providecommand{\DOIprefix}{doi:}
\providecommand{\ArXivprefix}{arXiv:}
\providecommand{\URLprefix}{URL: }
\providecommand{\Pubmedprefix}{pmid:}
\providecommand{\doi}[1]{\href{http://dx.doi.org/#1}{\path{#1}}}
\providecommand{\Pubmed}[1]{\href{pmid:#1}{\path{#1}}}
\providecommand{\bibinfo}[2]{#2}
\ifx\xfnm\relax \def\xfnm[#1]{\unskip,\space#1}\fi
\bibitem[{Acar et~al.(2018)Acar, Aksu, Uluagac and
  Conti}]{Acar2018AsurveyOnHomomorphicEncryptionSchemes}
\bibinfo{author}{Acar, A.}, \bibinfo{author}{Aksu, H.},
  \bibinfo{author}{Uluagac, A.S.}, \bibinfo{author}{Conti, M.},
  \bibinfo{year}{2018}.
\newblock \bibinfo{title}{{A survey on homomorphic encryption schemes: Theory
  and implementation}}.
\newblock \bibinfo{journal}{ACM Comput. Surv.} \bibinfo{volume}{51},
  \bibinfo{pages}{1--35}.
\bibitem[{Aslett et~al.(2015a)Aslett, Esperan~{\c {c}} a and
  Holmes}]{Aslett2015EncryptedStatisticalMachineLearning}
\bibinfo{author}{Aslett, L.J.M.}, \bibinfo{author}{Esperan~{\c {c}} a, P.M.},
  \bibinfo{author}{Holmes, C.C.}, \bibinfo{year}{2015}a.
\newblock \bibinfo{title}{{Encrypted statistical machine learning: new privacy
  preserving methods}}.
\newblock \bibinfo{journal}{arXiv:1508.06845v1} .
\bibitem[{Aslett et~al.(2015b)Aslett, Esperança and
  Holmes}]{Aslett2015ReviewHomomorphicEncryptionSoftware}
\bibinfo{author}{Aslett, L.J.M.}, \bibinfo{author}{Esperança, P.M.},
  \bibinfo{author}{Holmes, C.C.}, \bibinfo{year}{2015}b.
\newblock \bibinfo{title}{A review of homomorphic encryption and software tools
  for encrypted statistical machine learning}.
\bibitem[{Bost et~al.(2015)Bost, Popa, Tu and
  Goldwasser}]{Bost2015MachineLearningClassificationOverEncryptedData}
\bibinfo{author}{Bost, R.}, \bibinfo{author}{Popa, R.A.}, \bibinfo{author}{Tu,
  S.}, \bibinfo{author}{Goldwasser, S.}, \bibinfo{year}{2015}.
\newblock \bibinfo{title}{{Machine Learning Classification over Encrypted
  Data}}.
\bibitem[{Briguglio et~al.(2021)Briguglio, Moghaddam, Yousef, Traore and
  Mamun}]{Briguglio2021MachineLearningViaEncryption}
\bibinfo{author}{Briguglio, W.}, \bibinfo{author}{Moghaddam, P.},
  \bibinfo{author}{Yousef, W.A.}, \bibinfo{author}{Traore, I.},
  \bibinfo{author}{Mamun, M.}, \bibinfo{year}{2021}.
\newblock \bibinfo{title}{{Machine Learning via Encryption ({MLE}) Framework in
  Precision Medicine to Preserve Privacy}}.
\newblock
  \bibinfo{howpublished}{\url{https://github.com/isotlaboratory/Healthcare-Security-Analysis-MLE}}.
\bibitem[{Carey(2020)}]{Carey2020ExplanationImplementationThreeOpen}
\bibinfo{author}{Carey, A.}, \bibinfo{year}{2020}.
\newblock \bibinfo{title}{{On the Explanation and Implementation of Three
  Open-Source Fully Homomorphic Encryption Libraries Fully Homomorphic
  Encryption Libraries}}.
\bibitem[{Chakravarty et~al.(2017)Chakravarty, Gao, Phillips, Kundra, Zhang,
  Wang, Rudolph, Yaeger, Soumerai, Nissan, Chang, Chandarlapaty, Traina, Paik,
  Ho, Hantash, Grupe, Baxi, Callahan, Snyder, Chi, Danila, Gounder, Harding,
  Hellmann, Iyer, Janjigian, Kaley, Levine, Lowery, Omuro, Postow, Rathkopf,
  Shoushtari, Shukla, Voss, Paraiso, Zehir, Berger, Taylor, Saltz, Riely,
  Ladanyi, Hyman, Baselga, Sabbatini, Solit and
  Schultz}]{Chakravarty2017Oncokb}
\bibinfo{author}{Chakravarty, D.}, \bibinfo{author}{Gao, J.},
  \bibinfo{author}{Phillips, S.}, \bibinfo{author}{Kundra, R.},
  \bibinfo{author}{Zhang, H.}, \bibinfo{author}{Wang, J.},
  \bibinfo{author}{Rudolph, J.E.}, \bibinfo{author}{Yaeger, R.},
  \bibinfo{author}{Soumerai, T.}, \bibinfo{author}{Nissan, M.H.},
  \bibinfo{author}{Chang, M.T.}, \bibinfo{author}{Chandarlapaty, S.},
  \bibinfo{author}{Traina, T.A.}, \bibinfo{author}{Paik, P.K.},
  \bibinfo{author}{Ho, A.L.}, \bibinfo{author}{Hantash, F.M.},
  \bibinfo{author}{Grupe, A.}, \bibinfo{author}{Baxi, S.S.},
  \bibinfo{author}{Callahan, M.K.}, \bibinfo{author}{Snyder, A.},
  \bibinfo{author}{Chi, P.}, \bibinfo{author}{Danila, D.C.},
  \bibinfo{author}{Gounder, M.}, \bibinfo{author}{Harding, J.J.},
  \bibinfo{author}{Hellmann, M.D.}, \bibinfo{author}{Iyer, G.},
  \bibinfo{author}{Janjigian, Y.Y.}, \bibinfo{author}{Kaley, T.},
  \bibinfo{author}{Levine, D.A.}, \bibinfo{author}{Lowery, M.},
  \bibinfo{author}{Omuro, A.}, \bibinfo{author}{Postow, M.A.},
  \bibinfo{author}{Rathkopf, D.}, \bibinfo{author}{Shoushtari, A.N.},
  \bibinfo{author}{Shukla, N.}, \bibinfo{author}{Voss, M.H.},
  \bibinfo{author}{Paraiso, E.}, \bibinfo{author}{Zehir, A.},
  \bibinfo{author}{Berger, M.F.}, \bibinfo{author}{Taylor, B.S.},
  \bibinfo{author}{Saltz, L.B.}, \bibinfo{author}{Riely, G.J.},
  \bibinfo{author}{Ladanyi, M.}, \bibinfo{author}{Hyman, D.M.},
  \bibinfo{author}{Baselga, J.}, \bibinfo{author}{Sabbatini, P.},
  \bibinfo{author}{Solit, D.B.}, \bibinfo{author}{Schultz, N.},
  \bibinfo{year}{2017}.
\newblock \bibinfo{title}{Oncokb: A precision oncology knowledge base}.
\newblock \bibinfo{journal}{JCO Precision Oncology} , \bibinfo{pages}{1--16}.
\bibitem[{Dowlin et~al.(2016)Dowlin, Gilad-Bachrach, Laine, Lauter, Naehrig and
  Wernsing}]{Dowlin2016CryptoNetsApplyingNeuralNetworksToEncryptedData}
\bibinfo{author}{Dowlin, N.}, \bibinfo{author}{Gilad-Bachrach, R.},
  \bibinfo{author}{Laine, K.}, \bibinfo{author}{Lauter, K.},
  \bibinfo{author}{Naehrig, M.}, \bibinfo{author}{Wernsing, J.},
  \bibinfo{year}{2016}.
\newblock \bibinfo{title}{{Cryptonets: Applying neural networks to encrypted
  data with high throughput and accuracy}}, in: \bibinfo{booktitle}{33rd Int.
  Conf. Mach. Learn. ICML 2016}, pp. \bibinfo{pages}{342--351}.
\bibitem[{Gentry(2009)}]{Gentry2009FullyHomomorphicEncryptionScheme}
\bibinfo{author}{Gentry, C.}, \bibinfo{year}{2009}.
\newblock \bibinfo{title}{Fully homomorphic encryption scheme}.
\bibitem[{Graepel et~al.(2013a)Graepel, Lauter and
  Naehrig}]{Graepel2013Confidential}
\bibinfo{author}{Graepel, T.}, \bibinfo{author}{Lauter, K.},
  \bibinfo{author}{Naehrig, M.}, \bibinfo{year}{2013}a.
\newblock \bibinfo{title}{Ml confidential : machine learning on encrypted
  data}, in: \bibinfo{editor}{Kwon, T.}, \bibinfo{editor}{Lee, M.K.},
  \bibinfo{editor}{Kwon, D.} (Eds.), \bibinfo{booktitle}{Information Security
  and Cryptology - ICISC 2012 (15th International Conference, Seoul, Korea,
  November 28-30, 2012, Revised Selected Papers)},
  \bibinfo{publisher}{Springer}, \bibinfo{address}{Germany}. pp.
  \bibinfo{pages}{1--21}.
\newblock \bibinfo{note}{Conference; 15th International Conference on
  Information Security and Cryptography; 2012-11-28; 2012-11-30 ; Conference
  date: 28-11-2012 Through 30-11-2012}.
\bibitem[{Graepel et~al.(2013b)Graepel, Lauter and
  Naehrig}]{Graepel2013MlConfidentialMachineLearningOnEncryptedData}
\bibinfo{author}{Graepel, T.}, \bibinfo{author}{Lauter, K.},
  \bibinfo{author}{Naehrig, M.}, \bibinfo{year}{2013}b.
\newblock \bibinfo{title}{{ML confidential: Machine learning on encrypted
  data}}, in: \bibinfo{booktitle}{Lect. Notes Comput. Sci. (including Subser.
  Lect. Notes Artif. Intell. Lect. Notes Bioinformatics)}, pp.
  \bibinfo{pages}{1--21}.
\bibitem[{Hesamifard et~al.(2017)Hesamifard, Takabi and
  Ghasemi}]{Hesamifard2017DNNOverEncryptedData}
\bibinfo{author}{Hesamifard, E.}, \bibinfo{author}{Takabi, H.},
  \bibinfo{author}{Ghasemi, M.}, \bibinfo{year}{2017}.
\newblock \bibinfo{title}{{CryptoDL: Deep Neural Networks over Encrypted Data}}
  .
\bibitem[{Microsoft(2020)}]{Microsoft2019Github}
\bibinfo{author}{Microsoft}, \bibinfo{year}{2020}.
\newblock \bibinfo{title}{{M}icrosoft {SEAL} (release 3.6)}.
\newblock \bibinfo{howpublished}{\url{https://github.com/Microsoft/SEAL}}.
\newblock \bibinfo{note}{Microsoft Research, Redmond, WA.}
\bibitem[{Nandakumar(2019)}]{Nandakumar2019TowardsDNNTrainingOnEncryptedData}
\bibinfo{author}{Nandakumar, K.}, \bibinfo{year}{2019}.
\newblock \bibinfo{title}{{Towards Deep Neural Network Training on Encrypted
  Data}}.
\newblock \bibinfo{journal}{IEEE Conf. Comput. Vis. Pattern Recognit. Work.} .
\bibitem[{Penson et~al.(2020)Penson, Camacho, Zheng, Varghese, Al-Ahmadie,
  Razavi, Chandarlapaty, Vallejo, Vakiani, Gilewski
  et~al.}]{Penson2020DevelopmentGenomeDerivedTumor}
\bibinfo{author}{Penson, A.}, \bibinfo{author}{Camacho, N.},
  \bibinfo{author}{Zheng, Y.}, \bibinfo{author}{Varghese, A.M.},
  \bibinfo{author}{Al-Ahmadie, H.}, \bibinfo{author}{Razavi, P.},
  \bibinfo{author}{Chandarlapaty, S.}, \bibinfo{author}{Vallejo, C.E.},
  \bibinfo{author}{Vakiani, E.}, \bibinfo{author}{Gilewski, T.}, et~al.,
  \bibinfo{year}{2020}.
\newblock \bibinfo{title}{Development of genome-derived tumor type prediction
  to inform clinical cancer care}.
\newblock \bibinfo{journal}{JAMA oncology} \bibinfo{volume}{6},
  \bibinfo{pages}{84--91}.
\bibitem[{Rahman et~al.(2019)Rahman, Khalil, Alabdulatif and
  Yi}]{Rahman2019PrivacyPreservingServiceSelection}
\bibinfo{author}{Rahman, M.S.}, \bibinfo{author}{Khalil, I.},
  \bibinfo{author}{Alabdulatif, A.}, \bibinfo{author}{Yi, X.},
  \bibinfo{year}{2019}.
\newblock \bibinfo{title}{{Privacy preserving service selection using fully
  homomorphic encryption scheme on untrusted cloud service platform}}.
\newblock \bibinfo{journal}{Knowledge-Based Syst.} \bibinfo{volume}{180},
  \bibinfo{pages}{104--115}.
\bibitem[{Rivest et~al.(1978)Rivest, Adleman and
  Dertouzos}]{Rivest1978OnDataBanksAndPrivacyHomomorphism}
\bibinfo{author}{Rivest, R.}, \bibinfo{author}{Adleman, L.},
  \bibinfo{author}{Dertouzos, M.}, \bibinfo{year}{1978}.
\newblock \bibinfo{title}{{On Data Banks And Privacy Homomorphism}}.
\newblock \bibinfo{type}{Technical Report}. Massachusetts Institute of
  Technology.
\bibitem[{Sathya et~al.(2018)Sathya, Vepakomma, Raskar, Ramachandra and
  Bhattacharya}]{Sathya2018ReviewHomomorphicEncryptionLibraries}
\bibinfo{author}{Sathya, S.S.}, \bibinfo{author}{Vepakomma, P.},
  \bibinfo{author}{Raskar, R.}, \bibinfo{author}{Ramachandra, R.},
  \bibinfo{author}{Bhattacharya, S.}, \bibinfo{year}{2018}.
\newblock \bibinfo{title}{{A Review of Homomorphic Encryption Libraries for
  Secure Computation}} .
\bibitem[{Zehir et~al.(2017)Zehir, Benayed, Shah, Syed, Middha, Kim,
  Srinivasan, Gao, Chakravarty, Devlin
  et~al.}]{Zehir2017MutationalLandscapeMetastaticCancer}
\bibinfo{author}{Zehir, A.}, \bibinfo{author}{Benayed, R.},
  \bibinfo{author}{Shah, R.H.}, \bibinfo{author}{Syed, A.},
  \bibinfo{author}{Middha, S.}, \bibinfo{author}{Kim, H.R.},
  \bibinfo{author}{Srinivasan, P.}, \bibinfo{author}{Gao, J.},
  \bibinfo{author}{Chakravarty, D.}, \bibinfo{author}{Devlin, S.M.}, et~al.,
  \bibinfo{year}{2017}.
\newblock \bibinfo{title}{Mutational landscape of metastatic cancer revealed
  from prospective clinical sequencing of 10,000 patients}.
\newblock \bibinfo{journal}{Nature medicine} \bibinfo{volume}{23},
  \bibinfo{pages}{703}.

\end{thebibliography}

\clearpage

\section*{Supplementary Material}\label{sec:appendix}

\section{Genomics}\label{sec:genomics-1}
In this section we provide a more detailed description for the MSK-IMPACT
dataset~\citep{Zehir2017MutationalLandscapeMetastaticCancer}.

\subsection{Synopsis: MSK-IMPACT 2017}
\subsubsection*{Data\_CNA:} In this file, each column is a sample, and each row is a feature. Each
feature is represented by a string (e.g., EGFR) which is the \gls{HUGO symbol} for the corresponding
gene. The feature values are floats, which represent the \gls{copy number alteration} of the
feature’s gene. Positive and negative numbers represent duplication or deletion of repeat
nucleotides respectively.

\subsubsection*{Data\_Fusion:} This file structure is not as straightforward from an ML perspective.
Some general information is easily parsable, but more specific information contained in the
\texttt{comments} column requires more creative approaches and possibly expert (biochemical
background) consultation. The file comprises many rows, each with a gene HUGO symbol and sample ID,
among other fields (see below). Each row describes half a \gls{fusion} in which two genes are mixed
together via deletion, inversion, or translocation. Since two genes are involved, two rows describe
each half of the fusion effect. The only time this is not true is when the fusion is intergenic.
This is when a gene is mixed with DNA material which is located between genes, that is, is
non-coding. In this case only the information for the coding DNA material in the fusions is listed
in a single row. The remaining fields in each row are as follows:%
\begin{itemize}[partopsep=0in,parsep=0in,topsep=0.05in,itemsep=0.05in,leftmargin=0.15in]
  \item Huge\_Symbol: unique gene identified for the 410 genes sequenced with some being post-fixed
  with an Arabic numeral, a Latin letter, or in special cases both to indicate membership in a
  \gls{gene family}
  \item Entrez\_Gene\_ID: a gene integer ID used by the National Center for Biotechnology
Information (NCBI), only present for two samples
\item Center: where the sample was taken; set to ``MSKCC-DMP'', signifying the Memorial Sloan
Kettering Cancer Center, for all samples in the file
  \item Tumor\_Sample\_Barcode: sample ID
  \item Fusion: indicating which gene(s) are involved in the fusion
  \item DNA support: ``yes'' for all samples, indicating that fusion was detected with DNA sequencing
  \item RNA support: ``unknown'' for all samples, indicating that RNA-sequencing was not preformed
  \item Method: is ``NA'' for all samples
  \item Frame: either ``unknown'', ``\gls{out-of-frame}'', or ``\gls{in-frame}'' indicating the
  effect of each fusion
  \item Comments: short comments written by practitioners giving specifics of the \gls{mutation}
\end{itemize}

\subsubsection*{Data\_SV:} \gls{structural variation}, is a more general classification for
mutations than fusion mutations or copy number variations. There are 31 categorical, numerical, and
text features describing the type, location, and prevalence of the structural variances.%
\begin{itemize}[partopsep=0in,parsep=0in,topsep=0.05in,itemsep=0.05in,leftmargin=0.15in]
  \item Annotation: specific type and location of structural variation
  \item Breakpoint\_Type: the type of \gls{breakpoint}, that is, the junction between normal and
  rearranged
  \item Comments: small notes on some mutations
  \item Confidence Class: indicates the confidence in the final sequencing, and whether it was
automatically or manually determined
\end{itemize}
The remaining features are not documented in a single source but detail various aspects of the
structural variations such as sequencing method, the quality of sequencing, comparison with germline
DNA, variation location, and so on.

\subsubsection*{Data\_Mutation\_Significance\_Contribution:} This file contains 30 numerical
features corresponding to known \href{https://cancer.sanger.ac.uk/cosmic/signatures_v2}{mutation
  signatures}. For each sample, the feature value is the percentage of mutations explained by the
corresponding signature

\subsubsection*{Data\_Mutation\_Significance\_Confidence:} 30 numerical features that correspond to
the same \gls{mutation signatures} found abvove, but the feature value is the confidence in the
contribution scores instead. The mechanism used to determine confidence in contribution score is
described in Huang (2018)\footnote{\url{https://www.ncbi.nlm.nih.gov/pmc/articles/PMC5860213/}}

\subsubsection*{Data\_Mutation\_mskcc:} This file contains 46 columns describing mutations using a
subset of the columns found in the mutation annotation format (MAF) format. All columns and their
descriptions can be found in the GDC MAF
Format\footnote{\url{https://docs.gdc.cancer.gov/Data/File_Formats/MAF_Format/}}, except two:%
\begin{itemize}[partopsep=0in,parsep=0in,topsep=0.05in,itemsep=0.05in,leftmargin=0.15in]
  \item Hotspot: zero for all samples; was not configured when the table was made, so the same label
  was applied to all mutations
  \item cDNA\_change: the nucleotide change described with
  \href{https://www.ncbi.nlm.nih.gov/variation/hgvs/}{HGVS expression}
\end{itemize}

\subsubsection*{Data\_Mutation\_extended:} This is identical to Data\_Mutation\_mskcc but is missing
the ``cDNA\_change'' column

\subsubsection*{Data\_Gene\_Panael\_Matrix:} This file records which type of panel was used to
extract genomic data from the sample: the 341 gene or 410 gene panel. The 341 genes are a subset of
the 410 genes, so this distinction does not manifest in the other records because when creating a
record from a sample that used the 341 gene panel; the 69 genes are not tested for were marked as
not detected

\subsubsection*{Data\_cna\_hg19:} This file is the output of
\href{https://bioconductor.org/packages/release/bioc/html/DNAcopy.html}{DNAcopy}, which is an
open-source software package written in R that ``identifies genomic regions with abnormal copy
number'', that is, copy number alterations. Each row in the file corresponds to one such region or
``segment'' and has five columns describing it:%
\begin{itemize}[partopsep=0in,parsep=0in,topsep=0.05in,itemsep=0.05in,leftmargin=0.15in]
  \item \texttt{ID:} the sample ID
  \item \texttt{chrom:} the \gls{chromosome} the segment is in
  \item \texttt{loc.start:} start location of the segment
  \item \texttt{loc.end:} end location of the segment
  \item \texttt{num.mark:} number of probes bound to the segment
  \item \texttt{seg.mean:} the mean value, across all probes, of the segment; represents the
  \texttt{log2-ratio} of tumor copy number to normal copy number; a positive value indicates that
  the tumor has a higher copy number and vice versa
\end{itemize}

\subsubsection*{Data\_clinical\_patient:} Each row corresponds to a patient and contains five
columns:%
\begin{itemize}[partopsep=0in,parsep=0in,topsep=0.05in,itemsep=0.05in,leftmargin=0.15in]
  \item \#Patient Identifier: the patient ID
  \item Sex: the patient’s gender
  \item Patient's Vital Status: whether they are deceased or alive
  \item Smoking History: whether they previously, currently, or never smoked
  \item Overall Survival (Months): How long they survived since initial diagnosis; blank if they are
  currently alive or died after last follow up
  \item Overall Survival Status: the same as Patient Vital Status with a slightly different format
\end{itemize}

\subsubsection*{Data\_clinical\_sample:} Each row corresponds to a sample and contains 16 columns:%
\begin{itemize}[partopsep=0in,parsep=0in,topsep=0.05in,itemsep=0.05in,leftmargin=0.15in]
  \item \#Patient Identifier: the patient ID
  \item Sample Identifier: the sample ID
  \item Sample Collection Source: whether the sample was collected in house at MSKCC or by another
  party
  \item Specimen Preservation Type: method used to preserve the sample
  \item Specimen Type: how the specimen was collected
  \item DNA Input: amount of DNA in the sample in nanograms
  \item Sample Coverage: number of unique reads that included a given nucleotide during sequencing
  \item Tumor purity: percentage of cancer cells in sample
  \item Matched Status: whether the patient was matched with a gene therapy
  \item Sample Type: whether the sample came from the primary or metastatic tumor
  \item \Gls{primary site}: the location of the primary tumor
  \item Metastatic site: where the \gls{metastasis} occurred
  \item Sample Class: type of cancer tissue (is tumor for all samples)
  \item OncoTree Code: \href{http://oncotree.mskcc.org/#/home}{unique code} for specific tumor type
  \item Cancer Type: type of cancer that caused the tumor
  \item Cancer Type: Detailed: more specific sub-type of the cancer
\end{itemize}

\printnoidxglossary


\section{Homomorphic Encryption}\label{sec:encryption}
Below is an example presented by \cite{Sathya2018ReviewHomomorphicEncryptionLibraries} to introduce
the high-level concept of Homomorphic Encryption (HE).
\begin{enumerate}[partopsep=0in,parsep=0in,topsep=0.05in,itemsep=0.05in,leftmargin=0.2in]
  \item Let $m$ be the plain text message.
  \item Let a shared public key be a random odd integer $p$.
  \item Choose a random large $q$, small $r$,   $\left | r \right | \leq p\div 2 $.
  \item Ciphertext $c = pq + 2r +m$ (ciphertext $c$ is close to multiple of $p$).
  \item Perform homomorphic addition/multiplication as required.
  \item Decrypt: $m = (c \mod p) \mod 2$.
\end{enumerate}

Homomorphic addition can be illustrated as follows%
\begin{subequations}
  \begin{align}\label{eq:2}
    c_{1} &= q_{1}  \times  p + 2  \times  r_{1} +m_{1}\\
    c_{2} &= q_{2}  \times  p + 2  \times  r_{2} +m_{2}\\
    c_{1} + c_{2} &= (q_{1} + q_{2})  \times  p + 2  \times  (r_{1} + r_{2}) + (m_{1} + m_{2}),
  \end{align}
\end{subequations}
and Homomorphic multiplication as follows%
\begin{subequations}
  \begin{align}\label{eq:3}
    c1 &= q1  \times  p + 2  \times  r1 +m1\\
    c_{2} &= q_{2}  \times  p + 2  \times  r_{2} +m_{2}\\
    c_{1}  \times c_{2} &= ((c_{1}  \times q_{2}) +q_{1} \times c_{2}  \times q_{1}  \times q_{2}) \times p +\notag\\
       &2(2 \times r_{1}  \times r_{2}+r_{1}  \times m_{2}+m_{1}  \times r_{2})+m_{1}  \times m_{2}.
  \end{align}
\end{subequations}
Although homomorphic encryption holds massive potential in theory, it suffers from notable
shortcomings in practice. In many cases, it is limited to only addition and multiplication meaning
many functions must be approximated with high degree polynomials which incur a large computational
overhead. Even when using only this subset of operations, homomorphic operations are orders of
magnitude slower than conventional operations on plaintext data. Homomorphic encryption also leads
to substantial ciphertext expansion of a magnitude proportional to the targeted security strength.
Further, homomorphic encryption schemes do not allow unlimited operations without first decrypting
and re-encrypting or running an expensive denoising operation. These and other considerations make
homomorphic encryption not easily adaptable to practical applications without substantial foresight
and planning. For a more in depth review of the limitations and practical considerations of
homomorphic encryption~\citep{Aslett2015ReviewHomomorphicEncryptionSoftware}.


\end{document}